\documentclass[conference]{IEEEtran}
\IEEEoverridecommandlockouts
\usepackage{cite}
\usepackage{amsmath,amssymb,amsfonts,gensymb}
\usepackage{mathrsfs}
\usepackage{algorithmic}
\usepackage{graphicx}
\usepackage{textcomp}
\usepackage{xcolor}
\usepackage{bm}
\usepackage{subfigure}
\usepackage{hyperref}

\usepackage[]{changes}
\definechangesauthor[name={Juan Villacres}, color=blue]{JV}
\usepackage[numbers,sort&compress]{natbib}
\def\BibTeX{{\rm B\kern-.05em{\sc i\kern-.025em b}\kern-.08em
    T\kern-.1667em\lower.7ex\hbox{E}\kern-.125emX}}
\begin{document}

\title{Fusion-Driven Tree Reconstruction and Fruit Localization: Advancing Precision in Agriculture
	
\thanks{$^{1}$The author is with the Department of Electrical and Computer Engineering, University of California, Davis, Davis, CA 95616, USA}%
	
\thanks{$^{2}$The authors are with the Department of Biological and Agricultural Engineering, University of California, Davis, Davis, CA 95616, USA}%

\thanks{$^{3}$The author is with the Department of Mechanical and Aerospace Engineering, University of California, Davis, Davis, CA 95616, USA}%

\thanks{$^{4}$The author is with the Department of Plant Sciences, University of California, Davis, Davis, CA 95616, USA}%

\thanks{*These authors contributed equally}

}

\author{Kaiming Fu$^{*1}$, Peng Wei$^{*2}$,  Juan Villacres$^{2}$, Zhaodan Kong$^{3}$, Stavros G. Vougioukas$^{2}$, and Brian N. Bailey$^{4}$
}

\maketitle
\begin{abstract}

Fruit distribution is pivotal in shaping the future of both agriculture and agricultural robotics, paving the way for a streamlined supply chain. This study introduces an innovative methodology that harnesses the synergy of RGB imagery, LiDAR, and IMU data, to achieve intricate tree reconstructions and the pinpoint localization of fruits. Such integration not only offers insights into the fruit distribution, which enhances the precision of guidance for agricultural robotics and automation systems, but also sets the stage for simulating synthetic fruit patterns across varied tree architectures. To validate this approach, experiments have been carried out in both a controlled environment and an actual peach orchard. The results underscore the robustness and efficacy of this fusion-driven methodology, highlighting its potential as a transformative tool for future agricultural robotics and precision farming.

\end{abstract}

\section{Extended Summary}
\subsection{Introduction}

Understanding the spatial distribution of fruits within tree canopies is important in modern agriculture. Gaining insights into the precise arrangement of fruits does more than provide a count; it enhances harvest predictions, allowing for meticulous planning and efficient resource allocation. Moreover, as the trend of automation strengthens within agriculture, robots designed for tasks such as fruit picking rely heavily on this spatial data. When equipped with detailed knowledge of fruit locations, these robots can optimize their operations, reducing the risk of fruit damage, and ensuring streamlined movements through orchards \cite{wan2022real}.

Despite the pressing need for accurate 3D fruit distribution data, current tools and methodologies present notable challenges. Vision-based methods, particularly those using RGB cameras, help in detecting fruit distribution using object detection and segmentation techniques, allowing them to identify and count fruits in images~\cite{koirala2019deep}. However, challenges such as low resolutions and occlusions from nearby fruits or foliage can impact their accuracy. The adoption of high-resolution LiDAR technology is noteworthy for its capability to generate intricate 3D point clouds, capturing the complex structure of trees and fruit canopies. This offers a promising solution to the limitations of camera-based methods. Yet, LiDAR devices come with their own set of hurdles. The cost of high-resolution LiDAR can be prohibitive for many in the agricultural sector \cite{hammerle2014effects}. Additionally, many of these devices rely on GPS signals for data alignment, primarily when registering multiple scans from different locations. In GPS-denied environments such as dense orchards or beneath thick canopies, this becomes a significant issue. Furthermore, the stationary nature of these heavy LiDAR devices can lead to data blind spots, especially for fruits obscured from a fixed viewpoint. Efforts to augment LiDAR with RGB imagery have been made, aiming to enrich the point clouds with color and texture.

\textit{Contributions:} In this work, we present a solution to the identified challenges with our innovative handheld device and sensor fusion algorithm. The device, equipped with an inertial measurement unit, an RGB camera, and a LiDAR, is adept at capturing the 3D intricacies of trees and their fruit distributions, offering several compelling advantages: 1) it is cost-effective which ensures broader accessibility; 2) it leverages Simultaneous Localization And Mapping (SLAM) technology to reconstruct the environment and determine the location without depending on GPS, ensuring both robustness in operation and consistency in data capture; 3) the handheld design of our device allows that it can be maneuvered to various angles and positions, offering flexibility that stationary systems lack and yielding a more complete dataset. Obtaining the 3D positions of fruits from point clouds involves selecting points corresponding to the fruits, a process achievable through either manual or automatic methods. By precisely mapping the spatial arrangement of fruits, robotic systems can optimize harvesting strategies with unparalleled precision. This information enables robots to selectively harvest ripe fruits while leaving others to mature, minimizing waste and enhancing overall yield quality. Another good application could be using the data to train a Neural Network, which is designed to enhance synthetic fruit distributions across diverse arrays of tree branches, encompassing both authentic and artificially generated trees.

\begin{figure}[t!]
	\centering
    \subfigure[]{
    \includegraphics[width = 3.4in]{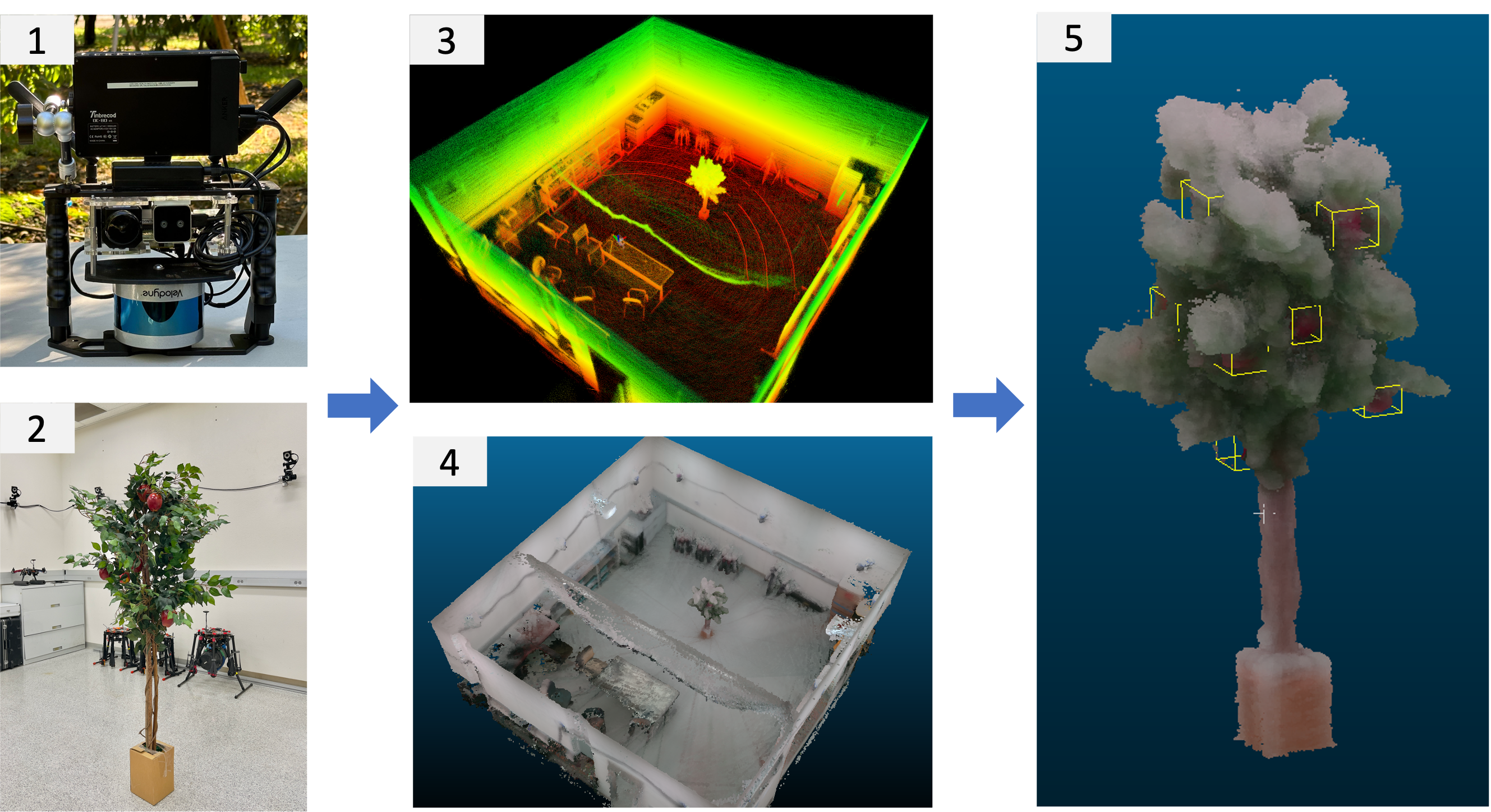}
    \label{fig:In_Lab_Experiment}	
    }
    \subfigure[]{
    \includegraphics[width = 3.4in]{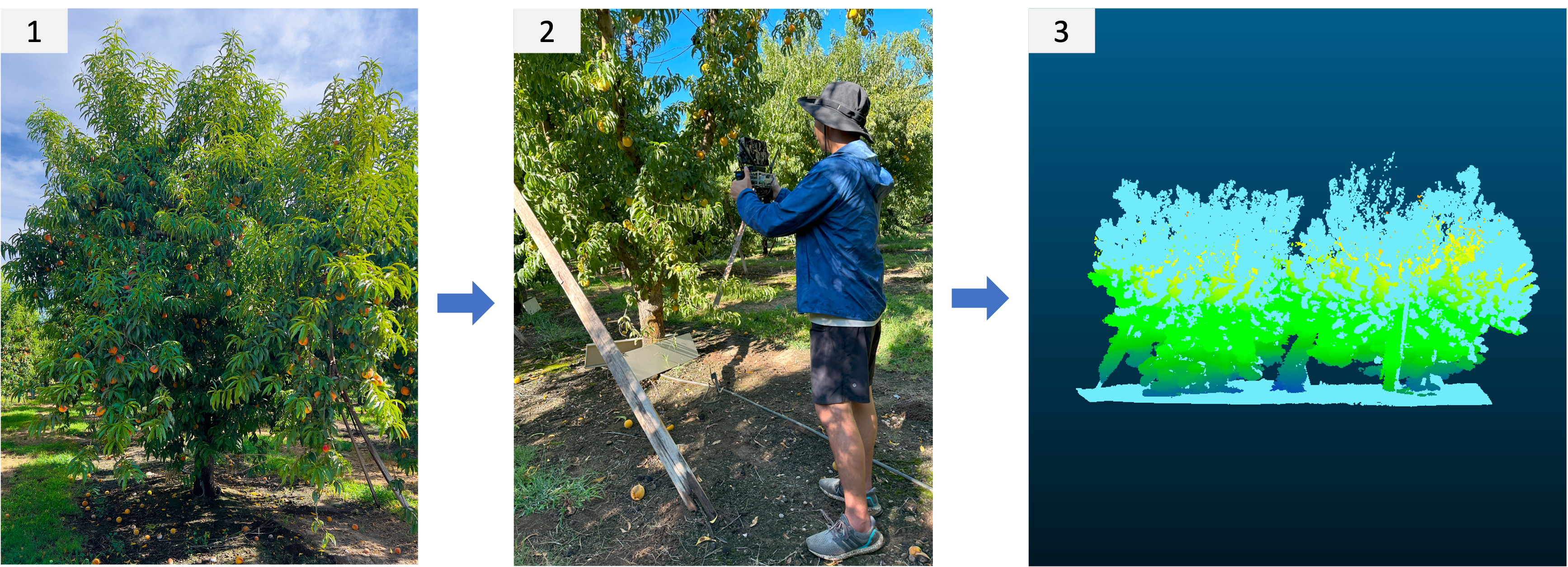}
    \label{fig:In_Field_Experiment}
    }
    \caption{(a) The workflow of the system and the results  obtained from an artificial apple tree in a controlled environment; (b) The scanning results of peach trees in a real peach orchard at Yuba City, CA.}
\end{figure}

\subsection{Methodology}
Figure~\ref{fig:In_Lab_Experiment} displays the handheld device we developed. The hardware includes a VectorNav VN-100 IMU, a Velodyne Puck Lite LiDAR, a Realsense D405 camera, a FLIR Vue thermal camera (not used), and an onboard Jetson TX2 computer. The intrinsic and extrinsic parameters of the sensors were calibrated carefully. The software pipeline utilizes a LiDAR-Inertial Odometry (LIO) SLAM algorithm~\cite{xu2022fast} to construct the geometric structure of the map. The map's texture is rendered through the RGB images using the odometry and camera information, resulting in a dense, 3D RGB-colored point cloud map of the surrounding environment. The pipeline was developed in the Robot Operating System (ROS). Subsequently, the fruit detection and localization can be done manually with human labeling or by using advanced computer vision technologies such as a re-trained YOLOv5 network and projecting its location from 2D to 3D \cite{francies2022robust}. 

\subsection{Experiments and Preliminary Results}
We devised two distinct experiments to rigorously evaluate the efficacy of our device and algorithm. The first experiment took place indoors, utilizing an artificial apple tree, while the second experiment was conducted in an actual peach field. The primary purpose of the indoor experiment was to assess the performance of our system in controlled conditions. In this setup, we meticulously positioned artificial apples within the tree's foliage to simulate real-world distribution. Conversely, the second experiment evaluated the device's performance in a more complex and dynamic environment. We deployed the device in a real peach field to determine its capability, where natural factors such as varying lighting conditions, foliage density, and occlusions were present. The findings and insights derived from our comprehensive analysis are presented in this paper, highlighting the outcomes of our study.

The results from the indoor experiment conducted on the artificial apple tree are illustrated in Fig.~\ref{fig:In_Lab_Experiment}. We successfully reconstructed the trees by fusing the data collected from different sensors. Fruit labeling was carried out manually, enabling us to identify the fruit positions within the point clouds. The ongoing development of the 3D fruit detection tool is aimed at automating the process of fruit position determination and recording. In the field experiment (see Fig.~\ref{fig:In_Field_Experiment}), we performed a comparative assessment between our handheld system and an established RIEGL VZ-1000 high-resolution LiDAR system in a commercial peach orchard, focusing on the scanning results in the real world. We scanned the trees using our device from different angles and locations to reconstruct the trees and then compared its result with the one obtained from the RIEGL VZ-1000 LiDAR. To better register the two resulting point clouds, we placed landmarks on each tree trunk. The registration was assessed using the average ratio (AR) metric~\cite{berens2021evaluation}, which takes into account the contribution of both point clouds and allows for consideration of multiple distance thresholds $d$. The threshold $d$ represents the minimum distance required for a point to be considered correctly registered. We obtained an AR of 0.89 (1 means a perfect match) using thresholds of 0.1, 0.5, 1, and 2.

\subsection{Future Work}

\bibliography{reference}

\end{document}